\documentclass[conference]{IEEEtran}
\IEEEoverridecommandlockouts
\usepackage{cite}
\usepackage{amsmath,amssymb,amsfonts}
\usepackage{graphicx}
\usepackage{textcomp}
\usepackage{xcolor}

\usepackage{xspace}
\usepackage{makecell}
\usepackage{booktabs,siunitx}
\usepackage{diagbox}
\usepackage{array, multirow}

\usepackage{algpseudocode}
\usepackage{geometry}
\usepackage{algorithm}

\def\BibTeX{{\rm B\kern-.05em{\sc i\kern-.025em b}\kern-.08em
    T\kern-.1667em\lower.7ex\hbox{E}\kern-.125emX}}
\begin{document}

\title{Are Vision Transformer Representations Semantically Meaningful? A Case Study in Medical Imaging\\

\author{Montasir Shams, Chashi Mahiul Islam, Shaeke Salman, Phat Tran, Xiuwen Liu % <-this % stops a space
%\thanks{*Denotes equal contributions}% <-this % stops a space
\thanks{ All authors are with the Department of Computer
Science, Florida State University, Tallahassee, Florida 32306, USA. Email:
{\tt\small
mshams@fsu.edu,ci20l@fsu.edu, salman@cs.fsu.edu, ttran@fsu.edu, liux@cs.fsu.edu, }%
}
}
}

\maketitle

\begin{abstract}
Vision transformers (ViTs) have rapidly gained prominence in medical imaging tasks such as disease classification, segmentation, and detection due to their superior accuracy compared to conventional deep learning models. However, due to their size and complex interactions via the self-attention mechanism, they are not well understood. In particular, it is unclear whether the representations produced by such models are semantically meaningful. In this paper, using a projected gradient-based algorithm, we show that 
their representations are not semantically meaningful and they are inherently vulnerable to small changes. Images with imperceptible differences can have very different representations; on the other hand, images that should belong to different semantic classes can have nearly identical representations. 
Such vulnerability can lead to unreliable classification
results; for example, unnoticeable changes cause the classification accuracy to be reduced by over 60\%. %. 
To the best of our knowledge, this is the first work to systematically demonstrate this fundamental lack of semantic meaningfulness in ViT representations for medical image classification, revealing a critical challenge for their deployment in safety-critical systems. 
\end{abstract}

\section{Introduction}
\label{sec:introduction}
% First section, background and importance
% The rapid advancements in large pre-trained foundation models~\cite{bommasani2022} have revolutionized a wide range of tasks, achieving state-of-the-art performance on benchmark datasets, excelling in standardized tests, and even surpassing human experts in professional exams~\cite{openai2023gpt4,Law2023Chatgpt}. 

%While many applications have been and are being built on top of these models, 
%new writing
The rapid evolution of large pre-trained foundation models has reshaped the landscape of artificial intelligence across multiple disciplines. Foundation models, built upon massive amounts of data and extensive pre-training, have demonstrated unprecedented generalization capabilities across domains such as natural language processing, computer vision, and multimodal reasoning tasks\cite{bommasani2022,openai2023gpt4,Law2023Chatgpt,Protein2022Brandes,Usmle2023Kung}.
These models rely on a relatively small application-specific component, which is fine-tuned on top of the shared foundation model. At the core of these models are their representations or embeddings, which capture high-level features of the input data. Such models, however, are not well understood due to their complexity, even though the need for understanding and the risks of lacking are widely recognized and acknowledged. For example, transformers have become a hallmark component in models for diverse applications~\cite{vaswani2023attention,dosovitskiy2021image,devlin2019bert},
including medical imaging tasks like disease classification, segmentation, and detection ~\cite{SHAMSHAD2023102802}. Their ability to model long-range dependencies and capture global contextual information via the self-attention mechanism has led to consistent improvements over traditional convolutional neural networks (CNNs) in several medical imaging benchmarks\cite{deininger2022comparativestudyvisiontransformers,10.1007/978-3-030-87237-3_5,Manzari_2023}.

Although vision transformers have achieved impressive success in medical image classification, detection, and segmentation, a crucial question remains: \textit{Do the internal representations produced by vision transformers correspond to semantically meaningful clinical concepts?} 
% In other words, can we trust that the embeddings produced by these models faithfully capture the relevant medical features that human experts rely upon for decision-making? 
The answer to this question holds profound implications for clinical adoption, where interpretability, robustness, and trustworthiness are paramount. In clinical practice, a semantically meaningful representation should exhibit two key properties: stability (similar medical conditions should have similar representations) and distinctiveness (different medical conditions should have distinguishable representations). In medical diagnosis, practitioners rely on specific visual patterns and features to identify conditions. For example, in diabetic retinopathy grading, doctors look for specific patterns of blood vessel changes and lesions. If transformer representations are semantically meaningful, they should organize images based on these clinically relevant features rather than superficial visual similarities.
While numerous studies have focused on evaluating the classification performance of ViTs in medical imaging\cite{khan2023recentsurveyvisiontransformers,SHAMSHAD2023102802,10733732,deininger2022comparativestudyvisiontransformers}, relatively little attention has been paid to systematically evaluating whether their learned representations satisfy these desirable semantic properties. This gap is particularly concerning in safety-critical domains such as healthcare, where unreliable representations may lead to incorrect predictions with severe consequences.

In this paper, we explore the semantic meaningfulness of vision transformer representations in medical imaging. Rather than merely evaluating model robustness at the level of classification outputs, we directly probe the structure of the representation space itself. Specifically, we introduce \textbf{Projected Representation Matching (PRM)} framework that enables controlled manipulation of an input image's embedding towards that of a target image, while strictly limiting pixel-level modifications to remain imperceptible to human observers. Through extensive experiments using multiple medical imaging datasets and vision transformer architectures\cite{10.1007/978-3-030-87237-3_5,Manzari_2023}, we provide compelling evidence that current transformer representations lack robust semantic grounding.

In summary, our main contributions are as follows:

\begin{itemize}
  \item We investigate the semantic meaningfulness of representations in vision transformers and demonstrate their sensitivity to subtle perturbations where a normal image can be changed to a different class even though the changes are imperceptible. For instance, the embedding of a normal image can be modified to match that of a diseased image, with imperceptible changes to the original image.
  %\item We develop a new algorithm that can detect the modifications reliably and efficiently.
  \item We demonstrate the results on multiple models (MIL-VT~\cite{10.1007/978-3-030-87237-3_5}, MedViT~\cite{Manzari_2023}) and datasets, providing empirical evidence of the representation dynamics. Our experiments reveal that the proposed approach is model-agnostic and dataset-agnostic.
  
  %\item We also propose a smoothness-based detection mechanism as a potential mitigation strategy for identifying manipulated inputs.
  %\item We show the implications of the findings to the current AI ecosystem built on these foundational models and propose ways to overcome them.
\end{itemize}

%Apart from the main contributions, we have performed several other things:
%\begin{itemize}
%\item Design a reliable procedure to compute the LDLC and we demonstrated the distribution in null space, subspace and manifolds. 
%\item We are also able to explore those spaces effectively using gradient descent, minimum norm, etc.

%\end{itemize}
\iffalse
The remainder of this paper is organized as follows: Section 2 reviews related work in medical imaging transformers and adversarial robustness. Section 3 describes the proposed PRM framework in detail. Section 4 presents comprehensive experimental results and analyses. Section 5 discusses potential mitigation strategies and broader implications, and finally, Section 6 concludes with directions for future research.
\fi

\section{Related Work}

The robustness and semantic reliability of representations in deep neural networks have attracted growing research attention across multiple domains. 
In this section, we review prior works related to (i) vision transformers in medical imaging, and  (ii) adversarial robustness of deep models.

\subsection{Vision Transformer in Medical Imaging}

%newly added

Transformers, originally designed for language tasks\cite{vaswani2023attention}, have demonstrated superior performance in computer vision and are now widely adopted in medical imaging. Vision Transformers (ViTs), which process images as sequences of patch embeddings via self-attention, have been applied to classification, segmentation, detection, synthesis, and clinical reporting tasks\cite{SHAMSHAD2023102802,dosovitskiy2021image}. ViT-based models such as MIL-VT\cite{10.1007/978-3-030-87237-3_5} and MedViT\cite{Manzari_2023} tailor these architectures for medical contexts. MIL-VT integrates multi-instance learning for weakly labeled data, while MedViT combines convolutional and transformer layers to enhance multi-scale feature learning. Additionally, large-scale pretraining on biomedical corpora has given rise to medical foundation models like BioMedCLIP and PMC-CLIP\cite{zhang2025biomedclipmultimodalbiomedicalfoundation,lin2023pmcclipcontrastivelanguageimagepretraining}. Despite strong task performance, most of these studies focus on end-to-end accuracy, with limited attention to the semantic structure of the learned representations. In clinical settings, however, explainability and semantic alignment are vital for interpretability, physician trust, and regulatory compliance.

\subsection{Adversarial Robustness in Medical Imaging}

It is well known that deep neural networks are susceptible to adversarial examples; small, imperceptible perturbations lead to drastic changes in model output\cite{goodfellow2015explaining,szegedy2014intriguing,madry2019deep}.
However, relatively few studies have been done to evaluate the effect of adversarial attacks on these transformer-based models in the medical field. Using the attack methods developed for
CNN models~\cite{Dong_2024,paschali2018generalizabilityvsrobustnessadversarial,doi:10.1126/science.aaw4399,MA2021107332}, 
recent studies~\cite{Laleh2022.03.15.484515,bhojanapalli2021understanding}
show that
ViTs exhibit greater resilience against typical attacks compared to CNN models
due to more structured latent spaces. 
While these findings are promising, they also highlight the need for a deeper understanding of vision transformer representations, particularly their semantic meaningfulness. 
%A semantically meaningful representation should organize images in a way that aligns with their underlying clinical features, such as disease severity or specific pathological patterns. 

In this paper, using a gradient-based optimization algorithm, we are able to explore the representations in the neighborhood of an image. More specifically,
we can modify the representation of an image to match that of another image with unnoticeable changes. The results show that images with imperceptible differences can have very different representations, and at the same time, very different images can have very similar representations. Therefore, the representations given by vision transformers may not be semantically meaningful. We demonstrate this using multiple medical imaging-centered models and datasets.

While prior work has focused on input-perturbation attacks that cause misclassification, our work probes a deeper, more fundamental issue: the integrity of the representation space itself. We show that even before the final classification head, the embeddings lack semantic coherence, a vulnerability that input-output attacks may not fully reveal.
% Furthermore, we observe that embedding matched modified images are much more sensitive to Gaussian noise than the original images. Based on this property, we also develop a new algorithm that can efficiently detect adversarial modifications. 

\section{Proposed Framework}

In this section, we present a formal description of the methodology employed to systematically investigate the semantic meaningfulness of vision transformer (ViT) representations in medical imaging.

\subsection{A Gradient-based Procedure for Representation Matching}
%Note that it is much more challenging to find inputs that would match the representation of a target input. 
%Here we develop an algorithm that

Generally, we model the representation given by a (deep) neural network (including a transformer) as a function $f: \mathbb{R}^m\rightarrow\mathbb{R}^n$.
We would like to analyze the representations given by vision transformers.
A key problem is to explore 
how images are mapped to representations.
As outlined in our earlier work\cite{salman2024intriguingequivalencestructuresembedding}, the proposed embedding alignment approach finds other representations in a given neighborhood of an image through matching. At the core of this method is an iterative gradient-based optimization process. Unlike standard neural network training, where gradients are computed for model parameters, this procedure computes gradients for the input itself.
More specifically, we define a loss for finding an input matching a given representation as
\begin{equation}
    L(x)=L(x_0+\Delta x)= \frac{1}{2}||f(x_0+\Delta x) - f(x_{tg})||^2,
\end{equation}
where $x_0$ is an initial input and $f(x_{tg})$ specifies the target embedding. The gradient of the loss function with respect to the input is given by:
\begin{equation}
\frac{\partial L}{\partial x} \approx \left(\frac{\partial f}{\partial x}|_{x=x_0}\right)^T(f(x_0+\Delta x) - f(x_{tg})).
\label{eq:grad_J}
\end{equation}

In each step, we calculate the loss as defined
by the loss function between the input image embedding
and the target image embedding. Then, using PyTorch, it
computes the gradient by doing backward computation. After
the gradient is calculated, we update the pixel values by doing
gradient descent.
Eq. \ref{eq:grad_J} shows how the gradient of the mean square loss function is related to the Jacobian of the representation function at $x=x_0$. 
While alternative solutions can be obtained,
by solving a quadratic programming problem or linear programming problem, depending on the norm to be used when minimizing $\Delta x$, we employ a gradient-descent approach that has proven effective across all our experiments. The success of this method is attributed to the structure of the transformer's Jacobian matrix. 

%Despite the presence of non-linear components such as softmax and ReLU, the vision transformer's behavior remains sufficiently locally linear, allowing the gradient-descent optimization to converge reliably.

% Due to this property, gradient-descent-based optimization is effective for all the cases we have tested due to the Jacobian of the transformer.

\begin{algorithm}[ht]
\caption{Projected Representation Matching (PRM) Algorithm}
\footnotesize % Reduce font size for algorithm content
\begin{algorithmic}[1]
\Require Source image $x_0$, target embedding $f(x_{\text{tg}})$, learning rate $\eta$, perturbation bound $\epsilon$, maximum iterations $T$
\Ensure Optimized perturbed image $\tilde{x}$
\State Initialize: $\tilde{x} \gets x_0$
\For{$t = 1$ to $T$}
    \State Compute loss: $L(\tilde{x}) = \frac{1}{2} \| f(\tilde{x}) - f(x_{\text{tg}}) \|^2$
    \State Compute gradient: $\nabla_{\tilde{x}} L$ via automatic differentiation
    \State Update perturbed image: $\tilde{x} \gets \tilde{x} - \eta \nabla_{\tilde{x}} L$
    \State Project perturbation back into allowed $\epsilon$-ball:
    \[
    \Delta x_{i,j,k} = \text{clip}\left( \tilde{x}_{i,j,k} - x_{0,i,j,k}, -\epsilon, \epsilon \right)
    \]
    \[
    \tilde{x}_{i,j,k} \gets x_{0,i,j,k} + \Delta x_{i,j,k}
    \]
    \If{convergence criterion met (e.g., $\| f(\tilde{x}) - f(x_{\text{tg}}) \|$ below threshold)}
        \State \textbf{break}
    \EndIf
\EndFor
\State \Return $\tilde{x}$
\end{algorithmic}
\label{alg:PRM}
\end{algorithm}

\subsection{Projections for Minimal Pixel Changes}
Another challenge for matching representations of medical images as a vulnerability identification method is ensuring that the semantics of the original image and the modified images are not altered to any medical experts. We achieve that using projections~\cite{madry2019deep}. In other words, a maximal allowed change to any pixel value is given as a hyperparameter. After each optimization step, any pixel value that would have been modified by the gradient descent algorithm more than the specified threshold will be clamped to the closest value within the allowed range. We denote the complete procedure as the  \textbf{\textit{projected representation matching algorithm (PRM)}}.

 \begin{figure}[ht]
  \centering
 % \vspace{-0.15in}
  %\subfloat[]
  {\includegraphics[width=0.5\columnwidth]{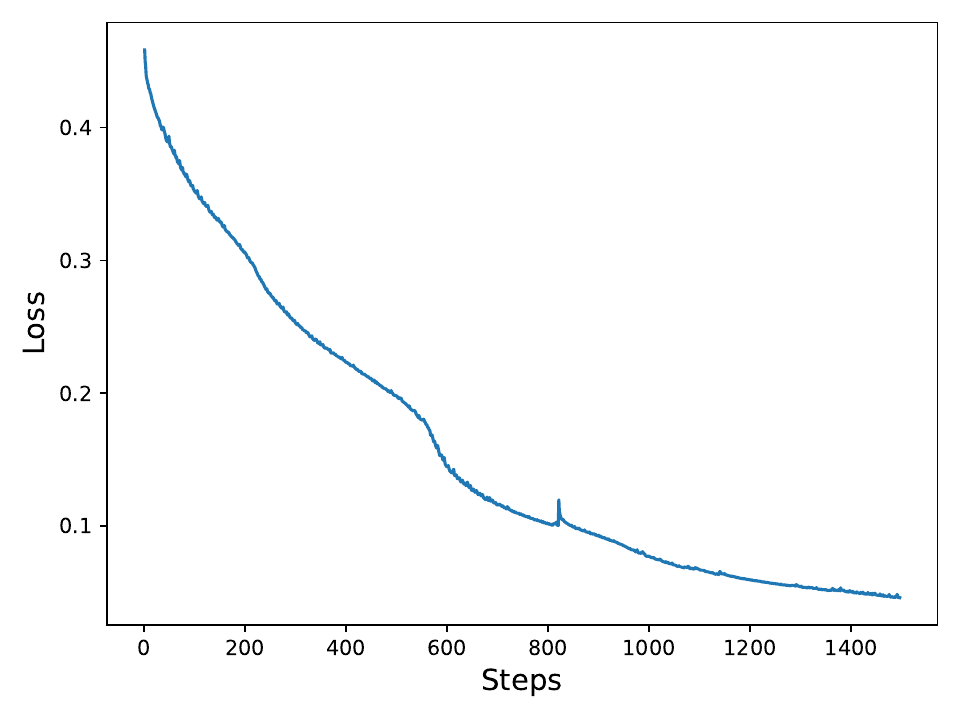}}%\label{fig:non1}}
  %\subfloat[]
  {\includegraphics[width=0.5\columnwidth]{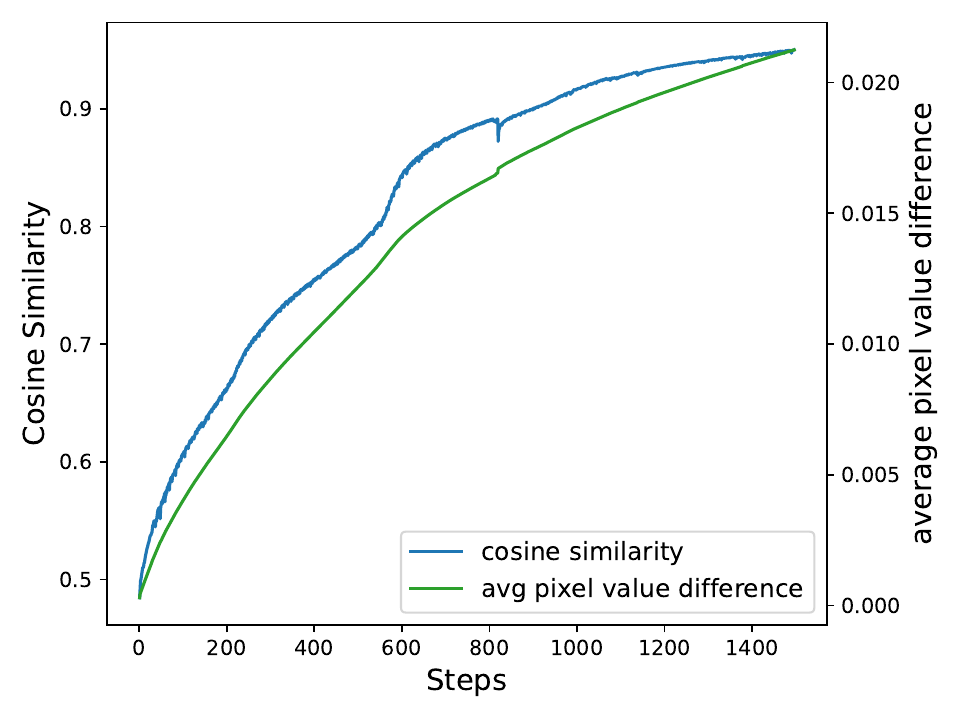}}%\label{fig:non2}}
  \vspace{-0.10in}
  \caption{The progression of loss in the process of aligning with a target embedding is depicted in the figure. (left), the graph illustrates the loss w.r.t steps. (right), the plot showcases the cosine similarity between the embeddings of the new input and the target across different steps. The average pixel value difference between the new input and the original image is also presented in the same graph for comprehensive insight.}
  \label{fig:training_dynamics}
  %\vspace{-0.15in}
\end{figure}

\section{Experiments and Results}
%In this section, we begin by outlining the experimental settings and implementation details. Our proposed method is systematically applied to various datasets. Subsequently, we delve into the outcomes of the experiments and provide quantitative results in the following subsections.

In this section, we present a comprehensive set of experiments designed to evaluate the semantic meaningfulness and vulnerability of vision transformer representations under the proposed Projected Representation Matching (PRM) framework. Our experiments are conducted across multiple medical imaging datasets and model architectures to assess both the generality and consistency of our findings.

\subsection{Experimental Setup}

\textbf{Datasets:} To thoroughly assess the effectiveness of our proposed method, we have primarily conducted experiments using two datasets: the APTOS2019 blindness dataset\footnote{Obtained from https://www.kaggle.com/code/ratthachat/aptos-eye-preprocessing-in-diabetic-retinopathy/input.} and the RFMiD2020 dataset\footnote{Obtained from https://ieee-dataport.org/open-access/retinal-fundus-multi-disease-image-dataset-rfmid.}. The first one includes  fundus images used for diabetic retinopathy (DR) grading with five label categories\footnote{Ranging from 
0 - No DR, 1 - Mild, 2 - Moderate, 3 - Severe, and 4 - Proliferative DR.}.

The RFMiD2020 dataset contains 1,900 fundus images, each labeled with a binary disease indicator: 0 for normal images and 1 for images depicting various pathologies.

\iffalse
\textbf{Model Architectures}

In our experiments, we mainly focus on the medical imaging focused MIL-VT\cite{10.1007/978-3-030-87237-3_5}, and the MedViT\cite{Manzari_2023} model. 

\textbf{MIL-VT Model:}

The Multiple Instance Learning Vision Transformer (MIL-VT)\cite{10.1007/978-3-030-87237-3_5}is a variant of ViT specifically adapted for weakly supervised learning in medical imaging. MIL-VT integrates patch-level aggregation mechanisms that are well-suited for handling localized lesions common in fundus images.

\textbf{MedViT Model:}

MedViT\cite{Manzari_2023} is a more recent vision transformer architecture optimized for generalized medical image classification across multiple domains. It incorporates architecture-level modifications that enhance its robustness on heterogeneous biomedical data.
\fi

\textbf{Implementation details:} 
To demonstrate the feasibility and impacts of the proposed method on large models, we have used a publicly available pre-trained MIL-VT model\footnote{Available from https://github.com/greentreeys/MIL-VT}. %For the experimental purpose, 
We finetuned the model for downstream applications. %It utilizes the pretrained vision (ViT-base-patch16-224 86M params)~\cite{ref_lncs7}. 
The input dimensions are $384\times384\times3$, with an embedding dimension of 384. To ensure a thorough evaluation of our method on both datasets, we partition the data consistently. Specifically, for fine-tuning the model, we randomly allocate 70\% of the data for training, 10\% for validation, and the remaining 20\% for testing purposes. Furthermore, from the test data, we randomly select a subset to create sample pairs on which to apply our PRM procedure. For each image within this subset, we choose another image with a different ground truth label on which to apply the PRM procedure. We perform all our experiments on a lab workstation featuring two NVIDIA A5000 GPUs.

\begin{figure*}[ht]
  \centering
  \vspace{-0.20in}
  {\includegraphics[width=0.75\textwidth]{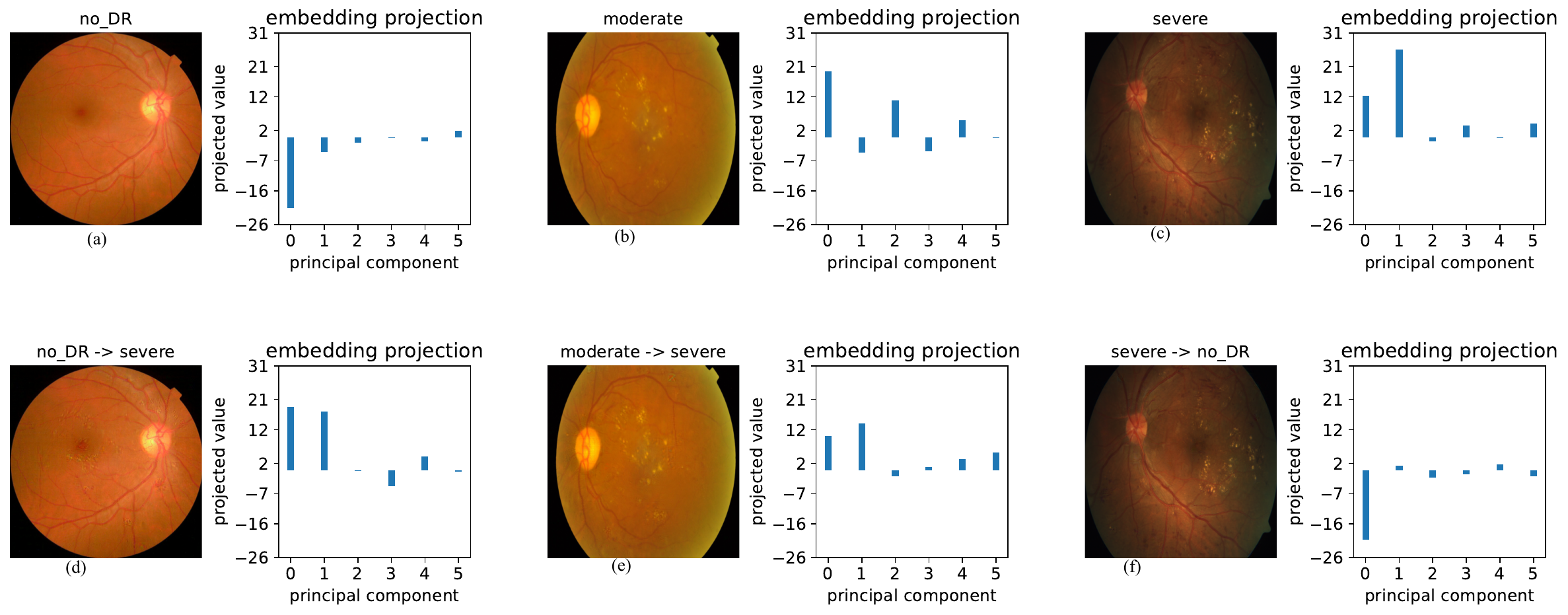}}%\label{fig:non1}}
  \vspace{-0.10in}
  \caption{Typical examples from the APTOS2019 obtained using the proposed framework. Three pairs of visually indistinguishable images ((a) and (d), (b) and (e), and (c) and (f)) have different representations, as shown in their low-dimensional projections. In contrast, very similar representations are seen for the images in (e), (c), and (f) despite their substantial semantic differences. Note that the arrow in the title ($original \rightarrow target$) signifies a derived image from the original one by aligning the embedding of the original image with the target image using our method. 
}
  \label{fig:overall}
  \vspace{-0.15in}
\end{figure*}

\subsection{Qualitative Evaluation}

We have conducted extensive evaluations  of the 
proposed %\textbf{projected representation matching procedure(
PRM algorithm using many different images. Fig. \ref{fig:training_dynamics} shows a typical example, depicting the evolution of loss during the matching process of a specified target embedding on the left side. On the right side, the figure demonstrates the steady increase in cosine similarity throughout the procedure. Notably, the algorithm exhibits robustness across a wide range of learning rates, varying from 0.001 to 0.9. For instance, convergence is achieved in approximately 8,000 iterations with a learning rate of 0.001, while only around 1500 iterations are required for a learning rate of 0.09. Importantly, visual inspection reveals negligible differences in the resulting images, underscoring the effectiveness and stability of the algorithm across various settings.

Fig. \ref{fig:overall} illustrates various images, their corresponding representations, and classification outcomes. Notably, Fig. \ref{fig:overall} reveals three pairs of visually indistinguishable images—(a) and (d), (b) and (e), and (c) and (f). Despite their visual similarities, these pairs exhibit significantly different representations, evident in their low-dimensional projections. Conversely, images in (e) and (c), despite stark semantic differences, display remarkably similar representations. For projections, we compute the top six principal components from a dataset subset and project embeddings onto them. These projections illustrate embedding differences, as similar embeddings yield similar projections, regardless of the specific components used.

\begin{table}[ht]
\vspace{-0.10in}
\begin{center}
\caption{Model performance under PRM with different maximum pixel changes ($\epsilon$). Learning rate (lr) is 0.9 for MIL-VT and 0.09 for MedViT-L experiments. Accuracy drops shown in parentheses.}
\small  % Reduce font size
\resizebox{\columnwidth}{!}{%
\begin{tabular}{|l|l|l|c|r|r|}

\hline
Dataset & Model & Type & $\epsilon$ & \makecell{Acc.\\(\%)} & \makecell{MSR\\(\%)} \\
\hline
\multirow{4}{*}{\makecell[l]{APTOS\\2019}}
& \multirow{4}{*}{MIL-VT} 
& MIL & 0.02 & $5.03_{(-76.3)}$ & 73.9 \\
& & MIL & 0.10 & $5.03_{(-76.3)}$ & 74.4 \\
& & ViT & 0.02 & $6.03_{(-75.3)}$ & 76.4 \\
& & ViT & 0.10 & $5.53_{(-75.8)}$ & 75.4 \\
\hline
\multirow{4}{*}{\makecell[l]{RFMiD\\2020}}
& \multirow{4}{*}{MIL-VT}
& MIL & 0.02 & $27.0_{(-60.2)}$ & 73.0 \\
& & MIL & 0.10 & $27.0_{(-60.2)}$ & 73.0 \\
& & ViT & 0.02 & $26.5_{(-60.7)}$ & 73.5 \\
& & ViT & 0.10 & $26.0_{(-61.2)}$ & 74.0 \\
\hline
bloodMNIST & MedViT-L & - & 0.10 & $17.8_{(-73.3)}$ & 85.5 \\
\hline
dermaMNIST & MedViT-L & - & 0.10 & $51.5_{(-22.8)}$ & 73.5 \\
\hline
\end{tabular}
}
\end{center}

\label{accuracy_success_rate_combined}
\end{table}

\vspace{-0.10in}

\subsection{Quantitative Evaluation}
\textbf{Impacts on  Classification Accuracy}

The MIL-VT model utilizes two distinct types of embeddings: the ViT embedding, representing the class token's features, and the MIL embedding, encompassing the aggregated embeddings of individual patches. We consider both embeddings to assess the effectiveness of our proposed \textbf{PRM} method. Table 1 illustrates the variations in model accuracy and match success rates (MSR) for a specific learning rate across different epsilon values(maximally allowed change in pixel values) on diverse datasets. 
Here, a match is successful if the modified image
is classified as having the same label as the ground truth label of the target representation. On the APTOS2019 dataset, the finetuned MIL-VT model attains an accuracy of 81.3\%. However, employing the ViT embedding for generating embedding aligned images results in a substantial decrease in accuracy to an average of 5.78\%, whereas for the MIL embedding, it diminishes to 5.03\%. Additionally, the match success rate on the ViT embedding averages 75.88\%, as opposed to 74.12\% on the MIL embedding. For the RFMiD2020 dataset, the finetuned model achieves an accuracy of 87.2\%, which is reduced to 27\% when subjected to attacks on the MIL embedding, with an associated match success rate of 73.00\%. Note that the match success rate is not 100\% even though we are able to match the representations closely in all the cases, primarily due to the errors introduced by the classifier. 

% Similar behavior is observed when considering the MedViT model, as well as the bloodMNIST and dermaMNIST datasets. 

 To demonstrate the generalizability of our approach, notably, we test the proposed method on the MedViT~\cite{Manzari_2023} model using the MedMNIST~\cite{Yang_2023} dataset. 
The results show that the observed behavior of embeddings—i.e., limited semantic coherence and high sensitivity to small perturbations- is consistent across these diverse models and datasets. This consistency highlights the model-agnostic and dataset-agnostic nature of our findings.

\begin{table}[ht]
  \centering
  \caption{The average PSNR value and SSIM index between the original and embedding-aligned images; the average is computed based on 1000 examples for each dataset, and the images are strictly randomly chosen from the datasets with no postselection.}
  \small
  \resizebox{\columnwidth}{!}{%
  \begin{tabular}{llll}
    \toprule
    \textbf{Image} & \textbf{Image Type} & \textbf{Mean PSNR} & \textbf{Mean SSIM} \\
    \midrule
    APTOS2019 & $Original \rightarrow Optimized$ & 42.55 dB & 0.972 \\
    APTOS2019 & $Target \rightarrow Optimized$ & 28.66 dB & 0.764 \\
    RFMiD2020 & $Original \rightarrow Optimized$ & 43.47 dB & 0.982 \\
    RFMiD2020 & $Target \rightarrow Optimized$ & 27.45 dB & 0.742 \\
    bloodMNIST & $Original \rightarrow Optimized$ & 55.53 dB & 0.997 \\
    bloodMNIST & $Target \rightarrow Optimized$ & 12.73 dB & 0.401 \\
    dermaMNIST & $Original \rightarrow Optimized$ & 53.05 dB & 0.995 \\
    dermaMNIST & $Target \rightarrow Optimized$ & 14.53 dB & 0.553 \\
    \bottomrule
  \end{tabular}
  }
  \vspace{0.20in}
  
  \label{tab:psnr_ssim}
\end{table}
\vspace{-0.10in}

\textbf{Image Quality Evaluation} 
Peak Signal-to-Noise Ratio (PSNR) and Structural Similarity Index (SSIM) are commonly used metrics to quantify the differences between the original and modified images~\cite{alain2010psnr,morales2019dehaze}. PSNR effectively measures the detailed quality of an image, whereas SSIM provides an intuitive assessment of its structural integrity. We present the average PSNR and SSIM values between the original (or target) and manipulated (i.e., embedding-aligned) images across several datasets under consideration in Table \ref{tab:psnr_ssim}. For the APTOS2019 dataset, when comparing original images with their optimized versions, we observe high-quality preservation (PSNR: 42.55 dB, SSIM: 0.972), indicating that our PRM algorithm maintains strong visual fidelity while modifying the representations. However, when comparing the original images with their target images, we see more substantial differences (PSNR: 28.66 dB, SSIM: 0.7638), which is expected as these images represent different medical conditions.
 These metrics indicate that the image quality does not significantly degrade with minimal distortion. Due to resource and time constraints, we restricted the results to 1000 examples for Table \ref{tab:psnr_ssim}. We followed the approach by Szegedy et al.~\cite{szegedy2014intriguing}, where they used a smaller set (64 images) from ImageNet when calculating the average distortion of adversarial examples.

\textbf{Semantic Manipulation }
To quantitatively validate the semantic manipulation capabilities of our approach, we analyze the cosine similarity between embeddings of different image pairs. Table. \ref{tab:avg_cos_sim} presents these results across datasets. For optimized images, we observe substantially higher cosine similarities with their target embeddings (0.84 for APTOS2019 and 0.77 for RFMiD2020) compared to their original embeddings (0.34 and 0.37 respectively). This significant difference demonstrates that our optimization successfully shifts the image representations toward their targets while maintaining visual dissimilarity, as confirmed by our image quality metrics in Table 2. These findings provide quantitative evidence that the representation space of vision transformers can be manipulated in semantically concerning ways, as visually similar images can have dramatically different embeddings.

\begin{table}[ht]
  \centering
  \small
  
  \caption{Average cosine similarity between original and optimized image embeddings, and between optimized and target embeddings, for several datasets.}
  
  \resizebox{\columnwidth}{!}{%
  \begin{tabular}{lll}
    \toprule
    \textbf{Dataset} & \textbf{Image Type} & \textbf{Average Cosine Similarity} \\
    \midrule
    APTOS2019 & Original & 0.34 \\
    APTOS2019 & Target & 0.84 \\
    RFMiD2020 & Original & 0.37 \\
    RFMiD2020 & Target & 0.77 \\
    bloodMNIST & Original & 0.20 \\
    bloodMNIST & Target & 0.93 \\
    dermaMNIST & Original & 0.28 \\
    dermaMNIST & Target & 0.87 \\
    \bottomrule
  \end{tabular}
  }
  \vspace{0.20in}
 
  \label{tab:avg_cos_sim}
\end{table}
\vspace{-0.10em}

\iffalse
\subsection{Model-Agnostic and Dataset-Agnostic Behaviour} Our key finding is that the semantic meaning of the embeddings produced by transformer models is fundamentally limited. Specifically, we observe that different inputs can share highly similar embeddings, while visually indistinguishable inputs may yield drastically different embeddings. This misalignment in the representation space raises questions about the semantic meaningfulness of these embeddings, particularly in high-stakes applications like medical imaging. To demonstrate the generalizability of our approach, we conduct experiments across multiple transformer models and datasets. Notably, we test the proposed method on the MedViT~\cite{Manzari_2023} model using the MedMNIST~\cite{Yang_2023} dataset. 
The results show that the observed behavior of embeddings—i.e., limited semantic coherence and high sensitivity to small perturbations; is consistent across these diverse models and datasets. This consistency highlights the model-agnostic and dataset-agnostic nature of our findings. 
%Please refer to the Supplemental Material for additional results with other models and datasets. 
\fi

\vspace{-0.15in}
\section{Potential Mitigation Technique} Since these limitations are inherent to the learned representations, conventional robustness training methods are generally insufficient to address classification changes; for instance, Mao et al. demonstrate that simply adjusting the final linear layer fails to resolve this issue~\cite{mao2023understanding}. A promising direction for practical mitigation involves incorporating smoothness-based detection mechanisms.

In our observations, embedding-matched images display significantly higher sensitivity to Gaussian noise compared to original images. Building on this insight, we propose a simple yet effective detection algorithm: Gaussian noise with a predefined standard deviation is applied to the input image, after which both the original and perturbed images are classified. If the predicted labels remain consistent, the image is considered unmodified; if they differ, the image is flagged as potentially modified. This approach demonstrates reliable performance across a broad range of noise levels. A similar strategy has been successfully employed by Islam et al., who applied smoothness-based detection to the CLIP-powered LM-Nav visual navigation system on the real-world RECON robot dataset to identify malicious modifications~\cite{islam2024malicious}.

\section{Discussion}

Through our comprehensive analysis, we have shown that both the MIL-VT and the standard ViT models exhibit vulnerability in their representation spaces. While the MIL-VT architecture shows marginally better robustness, both models experience comparable degradation in accuracy when subjected to representation manipulation.  This vulnerability stems from our ability to leverage the representation space to find minimal pixel-level changes that significantly alter the model's internal representations. Although the induced perturbations may appear unnoticeable to the human eye, the corresponding alterations in the embedding space are substantial, which can mislead the models' decision-making processes.

For medical image analysis applications, several kinds of attacks, including the common white-box and black-box attacks along with other variations, have been identified~\cite{Dong_2024}. 
In our study, we focus on representation vulnerability, which is a white-box attack. 
By understanding the vulnerabilities, 
the attack can be transferred to models using 
similar core architectures. Furthermore, the representation vulnerability is due to the nature of mapping from the input space to the representation space, and it applies to all models with a similar mapping structure.

The implications of these findings extend beyond just accuracy metrics. For medical image analysis applications, the vulnerability we have identified operates at the representation level, making it fundamentally different from traditional adversarial attacks. Although previous studies have focused on input-output relationships, our work reveals that the internal representation mechanism itself may not be capturing clinically relevant features in a semantically meaningful way.

Furthermore, our analysis shows that this vulnerability is inherent to the representation mapping structure rather than being dataset-specific or architecture-specific. The consistent patterns observed across different datasets and architectures suggest a fundamental limitation in how these models process and represent medical images.

\section{Conclusion and Future Work}
\iffalse
In this paper, we show that vision transformers used for medical image classification are inherently vulnerable to subtle adversarial attacks; unnoticeable changes can cause the representation that the classifier layers rely on to change significantly.
We have also developed a new algorithm to reliably detect adversarial modifications. 
As our framework is model-agnostic, it  can be used
to analyze other vision transformers, which is being
explored. 
Furthermore, we would like to characterize the mapping
from the input space to the representation space more systematically and efficiently through mathematical analyses, it is being investigated. 
\fi
%Before this fundamental limitation can be addressed, such models should not be used for critical applications.
In this paper, we present a comprehensive analysis of the semantic meaningfulness of vision transformer representations in medical image classification. Through our proposed projected representation matching (PRM) procedure, we have demonstrated that current transformer architectures, despite their impressive performance metrics, may not be creating truly semantic representations. The representation space may not effectively capture clinically relevant features, raising concerns about deployment in medical settings where model interpretability and reliability are paramount. %As a future work, we would like to characterize the mappingfrom the input space to the representation space more systematically and efficiently through mathematical analyses, it is being investigated.
As future work, we aim to mathematically characterize the mapping from input space to representation space by analyzing the local linearity and curvature of ViT models. Such analyses will provide deeper insight into the geometric properties of learned embeddings and guide the design of more robust, semantically faithful architectures for clinical AI.

%\appendices
\bibliographystyle{IEEEbib}
\bibliography{reference}

\begin{thebibliography}{10}

\bibitem{bommasani2022}
Rishi Bommasani, Drew~A. Hudson, Ehsan Adeli, Others, and et~al.,
\newblock ``On the opportunities and risks of foundation models,''
\newblock {\em CoRR}, 2022.

\bibitem{openai2023gpt4}
OpenAI,
\newblock ``Gpt-4 technical report,'' 2023.

\bibitem{Law2023Chatgpt}
Jonathan~H. Choi, Kristin~E. Hickman, Amy Monahan, and Daniel~B. Schwarcz,
\newblock ``Chat{GPT} goes to law school,''
\newblock {\em Journal of Legal Education (Forthcoming)}, 01 2023.

\bibitem{Protein2022Brandes}
Nadav Brandes, Dan Ofer, Yam Peleg, Nadav Rappoport, and Michal Linial,
\newblock ``{Protein{BERT}: a universal deep-learning model of protein sequence and function},''
\newblock {\em Bioinformatics}, vol. 38, no. 8, pp. 2102--2110, 02 2022.

\bibitem{Usmle2023Kung}
Tiffany~H. Kung, Morgan Cheatham, Arielle Medenilla, Czarina Sillos, Lorie De~Leon, Camille Elepa\~{n}o, Maria Madriaga, Rimel Aggabao, Giezel Diaz-Candido, James Maningo, and Victor Tseng,
\newblock ``Performance of {ChatGPT} on {USMLE}: Potential for {AI}-assisted medical education using large language models,''
\newblock {\em PLOS Digital Health}, 2023.

\bibitem{vaswani2023attention}
Ashish Vaswani, Noam Shazeer, Niki Parmar, Jakob Uszkoreit, Llion Jones, Aidan~N. Gomez, Lukasz Kaiser, and Illia Polosukhin,
\newblock ``Attention is all you need,''
\newblock in {\em Advances in Neural Information Processing Systems}, 2017.

\bibitem{dosovitskiy2021image}
Alexey Dosovitskiy, Lucas Beyer, Alexander Kolesnikov, Dirk Weissenborn, Xiaohua Zhai, Thomas Unterthiner, Mostafa Dehghani, Matthias Minderer, Georg Heigold, Sylvain Gelly, Jakob Uszkoreit, and Neil Houlsby,
\newblock ``An image is worth 16x16 words: Transformers for image recognition at scale,''
\newblock in {\em International Conference on Learning Representations}, 2021.

\bibitem{devlin2019bert}
Jacob Devlin, Ming-Wei Chang, Kenton Lee, and Kristina Toutanova,
\newblock ``Bert: Pre-training of deep bidirectional transformers for language understanding,'' 2019.

\bibitem{SHAMSHAD2023102802}
Fahad Shamshad, Salman Khan, Syed~Waqas Zamir, Muhammad~Haris Khan, Munawar Hayat, Fahad~Shahbaz Khan, and Huazhu Fu,
\newblock ``Transformers in medical imaging: A survey,''
\newblock {\em Medical Image Analysis}, vol. 88, pp. 102802, 2023.

\bibitem{deininger2022comparativestudyvisiontransformers}
Luca Deininger, Bernhard Stimpel, Anil Yuce, Samaneh Abbasi-Sureshjani, Simon Schönenberger, Paolo Ocampo, Konstanty Korski, and Fabien Gaire,
\newblock ``A comparative study between vision transformers and cnns in digital pathology,'' 2022.

\bibitem{10.1007/978-3-030-87237-3_5}
Shuang Yu, Kai Ma, Qi~Bi, Cheng Bian, Munan Ning, Nanjun He, Yuexiang Li, Hanruo Liu, and Yefeng Zheng,
\newblock ``Mil-vt: Multiple instance learning enhanced vision transformer for fundus image classification,''
\newblock in {\em Medical Image Computing and Computer Assisted Intervention – MICCAI 2021: 24th International Conference, Strasbourg, France, September 27 – October 1, 2021, Proceedings, Part VIII}, 2021, p. 45–54.

\bibitem{Manzari_2023}
Omid~Nejati Manzari, Hamid Ahmadabadi, Hossein Kashiani, Shahriar~B. Shokouhi, and Ahmad Ayatollahi,
\newblock ``Medvit: A robust vision transformer for generalized medical image classification,''
\newblock {\em Computers in Biology and Medicine}, vol. 157, pp. 106791, May 2023.

\bibitem{khan2023recentsurveyvisiontransformers}
Asifullah Khan, Zunaira Rauf, Abdul~Rehman Khan, Saima Rathore, Saddam~Hussain Khan, Najmus~Saher Shah, Umair Farooq, Hifsa Asif, Aqsa Asif, Umme Zahoora, Rafi~Ullah Khalil, Suleman Qamar, Umme~Hani Asif, Faiza~Babar Khan, Abdul Majid, and Jeonghwan Gwak,
\newblock ``A recent survey of vision transformers for medical image segmentation,'' 2023.

\bibitem{10733732}
Giovanni Lonia, Davide Ciraolo, Maria Fazio, Massimo Villari, and Antonio Celesti,
\newblock ``Comparing cnns and vits for medical image classification leveraging transfer learning,''
\newblock in {\em 2024 IEEE Symposium on Computers and Communications (ISCC)}, 2024, pp. 1--7.

\bibitem{zhang2025biomedclipmultimodalbiomedicalfoundation}
Sheng Zhang, Yanbo Xu, Naoto Usuyama, Hanwen Xu, Jaspreet Bagga, Robert Tinn, Sam Preston, Rajesh Rao, Mu~Wei, Naveen Valluri, Cliff Wong, Andrea Tupini, Yu~Wang, Matt Mazzola, Swadheen Shukla, Lars Liden, Jianfeng Gao, Angela Crabtree, Brian Piening, Carlo Bifulco, Matthew~P. Lungren, Tristan Naumann, Sheng Wang, and Hoifung Poon,
\newblock ``Biomedclip: a multimodal biomedical foundation model pretrained from fifteen million scientific image-text pairs,'' 2025.

\bibitem{lin2023pmcclipcontrastivelanguageimagepretraining}
Weixiong Lin, Ziheng Zhao, Xiaoman Zhang, Chaoyi Wu, Ya~Zhang, Yanfeng Wang, and Weidi Xie,
\newblock ``Pmc-clip: Contrastive language-image pre-training using biomedical documents,'' 2023.

\bibitem{goodfellow2015explaining}
Ian~J. Goodfellow, Jonathon Shlens, and Christian Szegedy,
\newblock ``Explaining and harnessing adversarial examples,''
\newblock in {\em International Conference on Learning Representations}, 2015.

\bibitem{szegedy2014intriguing}
Christian Szegedy, Wojciech Zaremba, Ilya Sutskever, Joan Bruna, Dumitru Erhan, Ian Goodfellow, and Rob Fergus,
\newblock ``Intriguing properties of neural networks,''
\newblock {\em CoRR}, 2014.

\bibitem{madry2019deep}
Aleksander Madry, Aleksandar Makelov, Ludwig Schmidt, Dimitris Tsipras, and Adrian Vladu,
\newblock ``Towards deep learning models resistant to adversarial attacks,'' 2019.

\bibitem{Dong_2024}
Junhao Dong, Junxi Chen, Xiaohua Xie, Jianhuang Lai, and Hao Chen,
\newblock ``Survey on adversarial attack and defense for medical image analysis: Methods and challenges,''
\newblock {\em ACM Computing Surveys}, vol. 57, no. 3, pp. 1–38, Nov. 2024.

\bibitem{paschali2018generalizabilityvsrobustnessadversarial}
Magdalini Paschali, Sailesh Conjeti, Fernando Navarro, and Nassir Navab,
\newblock ``Generalizability vs. robustness: Adversarial examples for medical imaging,'' 2018.

\bibitem{doi:10.1126/science.aaw4399}
Samuel~G. Finlayson, John~D. Bowers, Joichi Ito, Jonathan~L. Zittrain, Andrew~L. Beam, and Isaac~S. Kohane,
\newblock ``Adversarial attacks on medical machine learning,''
\newblock {\em Science}, vol. 363, no. 6433, pp. 1287--1289, 2019.

\bibitem{MA2021107332}
Xingjun Ma, Yuhao Niu, Lin Gu, Yisen Wang, Yitian Zhao, James Bailey, and Feng Lu,
\newblock ``Understanding adversarial attacks on deep learning based medical image analysis systems,''
\newblock {\em Pattern Recognition}, vol. 110, pp. 107332, 2021.

\bibitem{Laleh2022.03.15.484515}
Narmin~Ghaffari Laleh, Daniel Truhn, Gregory~Patrick Veldhuizen, Tianyu Han, Marko van Treeck, Roman~D. Buelow, Rupert Langer, Bastian Dislich, Peter Boor, Volkmar Schulz, and Jakob~Nikolas Kather,
\newblock ``Adversarial attacks and adversarial robustness in computational pathology,''
\newblock {\em Nature Communications}, vol. 13, pp. 5711, 2022.

\bibitem{bhojanapalli2021understanding}
Srinadh Bhojanapalli, Ayan Chakrabarti, Daniel Glasner, Daliang Li, Thomas Unterthiner, and Andreas Veit,
\newblock ``Understanding robustness of transformers for image classification,''
\newblock in {\em IEEE/CVF International Conference on Computer Vision (ICCV)}, 2021.

\bibitem{salman2024intriguingequivalencestructuresembedding}
Shaeke Salman, Md~Montasir~Bin Shams, and Xiuwen Liu,
\newblock ``Intriguing equivalence structures of the embedding space of vision transformers,'' 2024.

\bibitem{Yang_2023}
Jiancheng Yang, Rui Shi, Donglai Wei, Zequan Liu, Lin Zhao, Bilian Ke, Hanspeter Pfister, and Bingbing Ni,
\newblock ``Medmnist v2 - a large-scale lightweight benchmark for 2d and 3d biomedical image classification,''
\newblock {\em Scientific Data}, vol. 10, no. 1, Jan. 2023.

\bibitem{alain2010psnr}
Alain Horé and Djemel Ziou,
\newblock ``Image quality metrics: Psnr vs. ssim,''
\newblock in {\em 2010 20th International Conference on Pattern Recognition}, 2010, pp. 2366--2369.

\bibitem{morales2019dehaze}
Peter Morales, Tzofi Klinghoffer, and Seung~Jae Lee,
\newblock ``Feature forwarding for efficient single image dehazing,'' 2019.

\bibitem{mao2023understanding}
Chengzhi Mao, Scott Geng, Junfeng Yang, Xin Wang, and Carl Vondrick,
\newblock ``Understanding zero-shot adversarial robustness for large-scale models,'' 2023.

\bibitem{islam2024malicious}
Chashi~Mahiul Islam, Shaeke Salman, Montasir Shams, Xiuwen Liu, and Piyush Kumar,
\newblock ``Malicious path manipulations via exploitation of representation vulnerabilities of vision-language navigation systems,'' 2024.

\end{thebibliography}

\end{document}